\documentclass[card,numbers]{verisarxiv}
\usepackage{pifont}
\usepackage{float}      %
\usepackage{xurl}
\usetikzlibrary{positioning,arrows.meta,fit,backgrounds}

\newcommand{\bench}{TreasuryQA Bench}
\newcommand{\cmark}{\ding{51}}
\newcommand{\xmark}{\ding{55}}
\let\underscorechar\_
\renewcommand{\_}{\underscorechar\allowbreak}   %
\newtcolorbox{examplebox}[1][]{enhanced,sharp corners,boxrule=0.5pt,colframe=verisline,colback=white,
  left=9pt,right=9pt,top=7pt,bottom=7pt,before skip=6pt,after skip=6pt,fontupper=\footnotesize,
  before upper={\ifblank{#1}{}{{\verisinter\bfseries\fontsize{9}{11}\selectfont\color{verisink}#1}\par\vskip 3pt}}}

\newcommand{\FigFailures}{%
\begin{figure}[t]
\centering
\includegraphics[width=\linewidth]{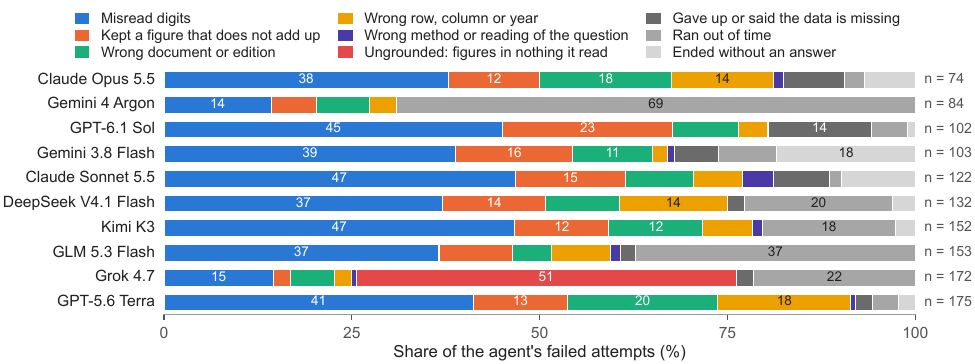}
\caption{\textbf{Misread digits are the most common failure for seven of the ten agents.} Bars split each agent's failed attempts ($n$; 1,269 in all) by primary cause; segments of 10\% or more are labelled.}
\label{fig:failures}
\end{figure}}

\newcommand{\FigTokens}{%
\begin{figure}[t]
\centering
\includegraphics[width=\linewidth]{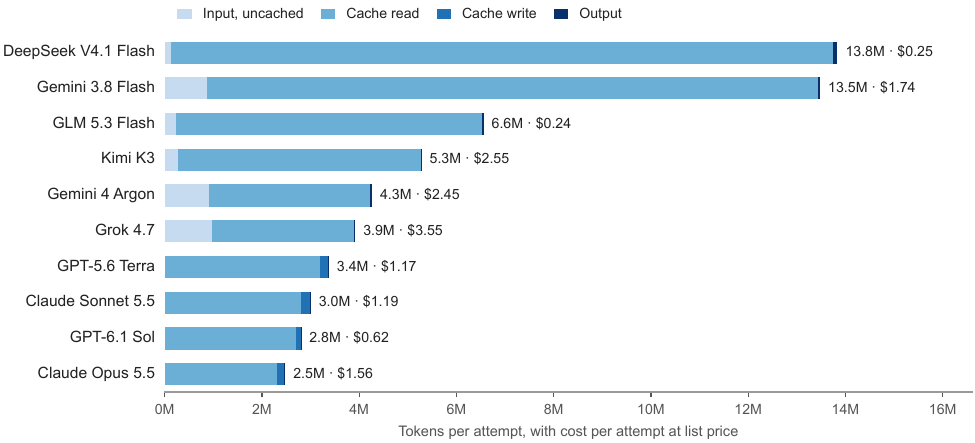}
\caption{\textbf{Tokens per attempt.} Mean tokens by kind for each agent, labelled with the total and the agent-side cost per attempt at list price.}
\label{fig:tokens}
\end{figure}}

\newcommand{\FigHeatmap}{%
\begin{figure}[t]
\centering
\includegraphics[width=\linewidth]{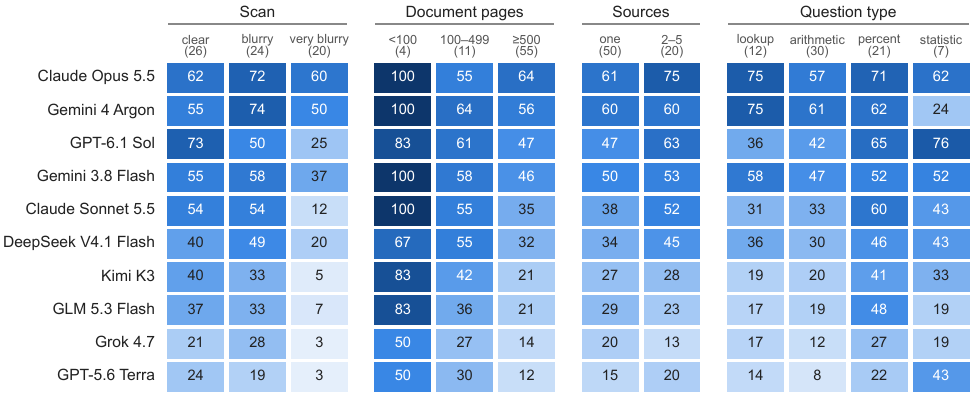}
\caption{\textbf{Very blurry scans separate the agents.} pass@1 (\%) by task tag over all three attempts; brackets give the number of tasks, and \emph{Sources} the documents a task draws on.}
\label{fig:heatmap}
\end{figure}}

\title[\bench]{\bench{}: Can AI Agents Work an Archive of Scanned Financial Records?}
\author{Ritiz Tambi}
\author{Sahar Shayegan}
\author{Joshua Meyer}
\author{Sun Kim}
\author{Ali Khan}
\author{Victor Shih}
\author{Mehdi Jamei}
\author{Andi Partovi}
\affiliation{Veris AI}
\kicker{Preprint}
\date{October 2026}
\correspondence{Ritiz Tambi at \href{mailto:ritiz@veris.ai}{ritiz@veris.ai}}
\abstract{Organisations keep decades of financial records as scanned PDFs on shared drives, and AI agents can now reach those drives through Model Context Protocol (MCP) servers. Existing benchmarks do not mirror this production setting because they simplify the task. In some, the relevant documents are handed to the agent, so it never has to search. In others, the agent searches through documents with pre-extracted text, so it never has to read scanned images. We introduce \bench{}, which requires both: 70 questions over 1,435 scanned U.S.\ Treasury publications, held in Veris's high-fidelity Dropbox twin. In each task the agent searches for the relevant file through Dropbox's MCP server, reads the pages it needs with Reducto's hosted OCR server or its own tools, and replies with one exact figure. A Python grader checks each answer against an accepted range around the correct value. In this paper we present results from ten agents. The best, Claude Opus 5.5 in Claude Code, solves 64.8\% of attempts and the worst, GPT-5.6 Terra in Codex, solves 16.7\%, so the benchmark is far from saturated. We also test how scan quality affects performance: every agent solves fewer tasks on the blurriest scans. Most failed attempts reply with a wrong number rather than no number, and the most common cause is a digit misread from the scan. Three worked examples are included with their answer keys.}

\begin{document}
\maketitle

\section{Introduction}
\label{sec:intro}

Suppose a human analyst is asked a question whose answer lies in decades of scanned financial records \citep{stanislawek2021kleister,opsahlong2026officeqapro}. First, the analyst must find the right file. Then they must locate the right table, read the right line and column, and check the figure against the table's totals. An AI agent given the same question goes through the same steps with tools: a cloud drive's Model Context Protocol (MCP) server to search for the file \citep{anthropic2024mcp,dropbox2026mcp}, and a hosted document reader that turns scanned pages into text \citep{reducto2026mcp}. The agent decides how to search the drive, which pages to read and whether to trust the text the reader returns.

Existing benchmarks only cover subparts of this workflow (Table~\ref{tab:comparison}, Section~\ref{sec:related}). Document QA benchmarks give the agent the page and might accept near misses \citep{mathew2021docvqa}; agentic and finance benchmarks give the agent pre-extracted, machine-readable text instead of page images; tool-use benchmarks grade the state an agent leaves in a service. OfficeQA Pro \citep{opsahlong2026officeqapro} asks for exact figures from scanned U.S.\ Treasury Bulletins, but its agents work on a local copy of the whole corpus on their own file system, rather than through the tools an agent would use in production.

\begin{table}[t]
\centering\footnotesize
\caption{\textbf{\bench{} and the closest benchmarks.} \emph{Scans}: page images without machine-readable text. \emph{Retrieval}: the agent must locate the source documents in a larger collection. \emph{Live tools}: tool servers or stateful services. $\circ$: partly (part of the corpus scanned; open web or a local copy only).}
\label{tab:comparison}
\setlength{\tabcolsep}{3.5pt}
\begin{tabular}{@{}llccclr@{}}
\toprule
Benchmark & What the agent reads & Scans & Retrieval & Live tools & Grading & Tasks \\
\midrule
DocVQA \citep{mathew2021docvqa} & one page image & \cmark & \xmark & \xmark & ANLS similarity & 50,000 \\
TAT-DQA \citep{zhu2022tatdqa} & 1--3 report pages & $\circ$ & \xmark & \xmark & exact match, F1 & 16,558 \\
GAIA \citep{mialon2024gaia} & web pages, files & \xmark & \cmark & $\circ$ & quasi-exact match & 466 \\
BrowseComp \citep{wei2025browsecomp} & open web & \xmark & \cmark & $\circ$ & LLM judge & 1,266 \\
OfficeQA Pro \citep{opsahlong2026officeqapro} & Treasury Bulletins & $\circ$ & \cmark & $\circ$ & exact match & 133 \\
Finance Agent \citep{bigeard2025financeagent} & SEC filings, web & \xmark & \cmark & \cmark & rubric, LLM judge & 537 \\
$\tau^2$-bench \citep{barres2025tau2} & service databases & -- & -- & \cmark & end state & 279 \\
MCPMark \citep{wu2025mcpmark} & MCP service state & -- & -- & \cmark & end-state scripts & 127 \\
\midrule
\bench{} (ours) & scanned volumes & \cmark & \cmark & \cmark & Python, range & 70 \\
\bottomrule
\end{tabular}
\end{table}

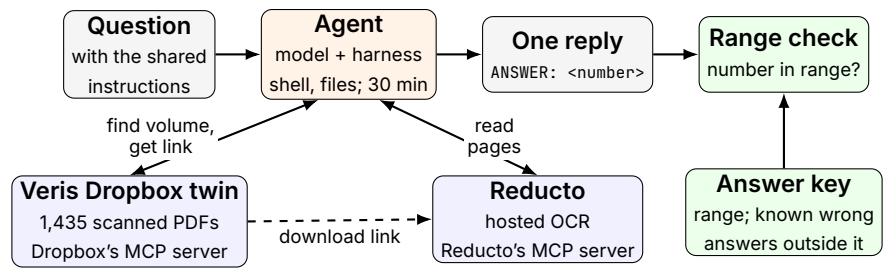
\begin{figure}[t]
\centering
\begin{tikzpicture}[font=\sffamily\small,
  box/.style={draw, rounded corners, align=center, inner xsep=3pt, inner ysep=2pt, minimum height=10mm},
  io/.style={box, fill=gray!8},
  agt/.style={box, fill=orange!10},
  svc/.style={box, fill=blue!6},
  grd/.style={box, fill=green!8},
  lab/.style={font=\sffamily\scriptsize, fill=white, inner sep=1pt, align=center},
  arr/.style={-{Latex[length=2mm]}, thick},
  biarr/.style={{Latex[length=2mm]}-{Latex[length=2mm]}, thick}]
\node[io] (q) {\textbf{Question}\\{\scriptsize with the shared}\\{\scriptsize instructions}};
\node[agt, right=6mm of q] (agent) {\textbf{Agent}\\{\scriptsize model + harness}\\{\scriptsize shell, files; 30 min}};
\node[io, right=6mm of agent] (reply) {\textbf{One reply}\\{\scriptsize\ttfamily ANSWER: <number>}};
\node[grd, right=6mm of reply] (check) {\textbf{Range check}\\{\scriptsize number in range?}};
\node[svc, below=10mm of agent, xshift=-29mm] (dbx)
  {\textbf{Veris Dropbox twin}\\{\scriptsize 1,435 scanned PDFs}\\{\scriptsize Dropbox's MCP server}};
\node[svc, below=10mm of agent, xshift=25mm] (red)
  {\textbf{Reducto}\\{\scriptsize hosted OCR}\\{\scriptsize Reducto's MCP server}};
\node[grd, below=10mm of check] (key)
  {\textbf{Answer key}\\{\scriptsize range; known wrong}\\{\scriptsize answers outside it}};
\draw[arr] (q) -- (agent);
\draw[arr] (agent) -- (reply);
\draw[arr] (reply) -- (check);
\draw[arr] (key) -- (check);
\draw[biarr] ([xshift=-4mm]agent.south) -- node[lab, left=2pt]{find volume,\\get link} (dbx.north);
\draw[biarr] ([xshift=4mm]agent.south) -- node[lab, right=2pt]{read\\pages} (red.north);
\draw[arr, dashed] (dbx.east) -- node[lab, below=2pt]{download link} (red.west);
\end{tikzpicture}
\caption{\textbf{One \bench{} attempt.} The agent finds the volume in a fresh copy of the Veris Dropbox twin through Dropbox's MCP server and hands a download link to Reducto's MCP server, which returns the pages it asks for. The attempt is solved if its \texttt{ANSWER:} number lies in the task's range.}
\label{fig:overview}
\end{figure}

We present \bench{} (Figure~\ref{fig:overview}). Its archive is the 1,435 image-only PDFs of the OfficeQA Pro v2 corpus \citep{databricks2026officeqaprov2}, U.S.\ Treasury publications from 1793 to 2024, held in Veris's high-fidelity Dropbox twin. Agents reach the documents through Dropbox's MCP server and read pages with Reducto's hosted MCP server; web search is off. Each of the 70 tasks is an analyst's question with one numeric answer, 55 written for \bench{} and 15 adapted from OfficeQA Pro v2. A Python grader checks each answer against an accepted range around the correct value (Section~\ref{sec:grading}).

We evaluate ten agents, each a model in its own command-line harness, three times per task (Table~\ref{tab:main}). Claude Opus 5.5 in Claude Code solves the most, 64.8\% of attempts, not significantly more than Gemini 4 Argon in Gemini CLI (60.0\%). The worst, GPT-5.6 Terra in Codex, solves 16.7\%, so the benchmark is far from saturated. Every agent solves fewer tasks on the blurriest scans. Most failed attempts reply with a wrong number rather than no number, most often because of a digit misread from the scan. Our contributions are:
\begin{itemize}[leftmargin=*,itemsep=0pt,topsep=1pt]
  \item \textbf{\bench{}}: 70 exact-figure questions over scanned Treasury records reached through Dropbox's MCP server, with answer keys computed by programs and checked against the scans (Section~\ref{sec:benchmark}).
  \item \textbf{A Python grader} that checks each answer against an accepted range (Section~\ref{sec:grading}).
  \item \textbf{An evaluation of ten model--harness pairs}, with cost, time, tool use and the causes of failure (Sections~\ref{sec:results} and~\ref{sec:analysis}).
\end{itemize}

\section{Related Work}
\label{sec:related}

\paragraph{Document understanding and OCR.} Scanned financial reports are already a benchmark subject: Kleister asks for amounts such as income and spending from charity financial reports, most of them scanned \citep{stanislawek2021kleister}. DocVQA scores answers about single page images by string similarity, so small OCR errors cost little \citep{mathew2021docvqa}. TAT-DQA asks questions, many needing arithmetic, about a few pages of financial reports and credits a number only if it is exact \citep{zhu2022tatdqa}. MMLongBench-Doc gives the model screenshots of every page and accepts decimal answers within 1\% \citep{ma2024mmlongbenchdoc}. OCR errors also carry through to the answers built on the extracted text \citep{zhang2025ohrbench}. All of these hand the system the document. In \bench{} the agent must first find the page in a large archive, and a 1\% tolerance would accept every wrong number given on the example task (Figure~\ref{fig:example}), most of them a faint 8 read as a 6.

\paragraph{Question answering over document collections.} GAIA \citep{mialon2024gaia} and BrowseComp \citep{wei2025browsecomp} ask short-answer questions over web pages and files, not archives of scanned documents. OfficeQA Pro \citep{opsahlong2026officeqapro} asks questions over U.S.\ Treasury Bulletins, graded by exact match; issues before 1996 are scans, and embedded text layers are removed. Its agents work on a local copy with OCR libraries installed, and giving them parsed text raises their accuracy. Its v2 release, a new 1793--2024 corpus that we use, ships parsed text with the PDFs \citep{databricks2026officeqaprov2}. AGORA asks numeric questions over workplace documents that agents explore in an offline sandbox \citep{guo2026agora}. \bench{} withholds the parsed text, puts the scans behind a file store's MCP server, gives every agent the same hosted reader, and sets each range against known wrong answers.

\paragraph{Financial question answering.} Financial QA benchmarks ask about modern corporate filings, such as earnings reports \citep{chen2021finqa} and full public filings \citep{islam2023financebench}, which come with machine-readable text. Agentic finance benchmarks have models research SEC filings, graded by an LLM judge with a rubric \citep{bigeard2025financeagent}. In \bench{} the documents exist only as page images, the agent must first find the right file in a drive, and each task has one numeric answer that a script can check.

\paragraph{Tool-use environments and harnesses.} Agent benchmarks increasingly grade the state an agent leaves in databases behind customer-service tools \citep{yao2024taubench,barres2025tau2} or in real MCP servers \citep{wu2025mcpmark}, and TheAgentCompany has agents work with files in a self-hosted drive, among other services \citep{xu2025theagentcompany}. \bench{} grades no final state. Its difficulty is in reading through Dropbox's and Reducto's tools. The harness around a model changes both accuracy and cost \citep{kapoor2026hal} and can reverse rankings \citep{li2026modelharness}. Each proprietary model therefore runs in its vendor's own CLI, which Terminal-Bench also pairs with that vendor's models \citep{merrill2026terminalbench}, and we name every agent by model and harness. We report cost with accuracy and follow the Agentic Benchmark Checklist \citep{zhu2025abc}.

\section{\bench{}}
\label{sec:benchmark}

\subsection{Design and Task Formulation}
\label{sec:formulation}
Three choices shaped \bench{}. The setting matches real use: scans with no machine-readable text in a cloud drive, reached through the drive's own MCP server. Every answer key is computed by a program from the source documents and confirmed on the scans. And a Python grader, not a model, decides whether an answer is right. The agent gets the question once, works alone, and ends with one reply. Figure~\ref{fig:example} shows an example task.

\begin{figure}[t]
\begin{examplebox}[Example task P092]
\textbf{Question.} ``As of June 30, 1945, what was the combined balance in the unemployment trust fund accounts of the States and Territories, excluding the railroad unemployment insurance account? Answer in dollars to the cent.''

\smallskip
\textbf{Where.} \nolinkurl{combined_statement__historical__cs-1945.pdf} (766 pages, 60\,ppi; the answer page is a 100\,ppi replacement), PDF page 668, Table~12, the Subtotal row, which excludes the railroad account; the question names neither.

\smallskip
\textbf{Solution.} The cell reads \$6,679,29?,407.45, the faint digit looking like a 6. Receipts minus withdrawals ($9{,}010{,}689{,}947.62 - 2{,}331{,}391{,}540.17$), Total minus the railroad account ($7{,}315{,}691{,}196.62 - 636{,}392{,}789.17$) and the sum of the 51 State and Territory balances all give 6,679,298,407.45. The oracle computes all three and asserts that they agree.

\smallskip
\textbf{Key.} \$6,679,298,407.45 $\pm$ \$1.00. \textbf{Known wrong answers} (5) include 6,679,296,407.45 (the faint 8 read as a 6), 7,315,691,196.62 (the Total, with the railroad account) and 6,209,588,494.49 (deposits minus withdrawals, without earnings).

\smallskip
\textbf{Outcome.} Solved in 11 of 30 attempts; 16 answered 6,679,296,407.45.
\end{examplebox}
\caption{\textbf{An example task.} The answer is one cell of one table. The agent must find the volume, pick the Subtotal row and settle a faint digit with the page's own totals.}
\label{fig:example}
\end{figure}

\subsection{Environment}
\label{sec:environment}
\paragraph{The archive.} The archive is the OfficeQA Pro v2 corpus \citep{databricks2026officeqaprov2}: 1,435 PDFs in one folder. We counted 118,524 pages and checked that none of the files has machine-readable text. Most are the annual \emph{Combined Statements of Receipts, Expenditures and Balances} (1872--2024); the rest are older reports on receipts and expenditures (1793--1893). Most files are scanned at 150\,ppi, but some are as coarse as 60\,ppi.

\paragraph{The Dropbox twin.} The archive sits in Veris's twin of Dropbox, built from recordings of the live service, and each attempt gets a fresh copy. Agents use Dropbox's MCP server with its real tool list and instructions \citep{dropbox2026mcp}, and we checked each behaviour below against the live service. \texttt{search} matches file names and file text, so for scans it matches names only. \texttt{get\_file\_content} returns a file's text, which is empty for a scan, and refuses files over 5\,MB. \texttt{download\_link} returns a short-lived link that agents pass to Reducto. The twin does not offer Dropbox's own OCR, so every agent reads pages through the same reader.

\paragraph{Reducto.} Every harness connects to Reducto's hosted MCP server \citep{reducto2026mcp}, which turns chosen pages of a document into text and tables. It reads documents from URLs, not from the agent's disk, so the agent passes a download link to \texttt{parse\_document}, or to \texttt{upload\_file}, which returns an identifier that later calls can reuse.

\paragraph{Harness, container and instructions.} Each attempt runs one harness in a fresh container with Python and common command-line tools, but no archive files and no OCR engine. Each harness keeps its built-in tools: a shell, file tools and, for Claude Code, sub-agents. Web search is off, but the shell can reach the network. The agent receives the shared instructions (Appendix~\ref{app:prompts}) and the question as one message; each CLI keeps its own system prompt. The instructions explain how files are named, tell the agent to find pages through a volume's contents and read small page ranges with Reducto, ask for the latest revision of a figure, as OfficeQA Pro does \citep{opsahlong2026officeqapro}, and require a final line \texttt{ANSWER: <final answer>}. Each attempt has 30 minutes.

\subsection{Task Construction}
\label{sec:construction}
We built the tasks in three steps: a script found candidate figures, LLM agents wrote a task around each, and scripts and a model solver tested the drafts. We then checked every task against the scans and reviewed it ourselves (Section~\ref{sec:qc}).

\paragraph{Finding candidate figures.} OfficeQA Pro v2 ships parsed text with the PDFs. Agents never see it, but we used it to build the tasks. A script searched its tables for cells whose value is fixed by the table's own row or column totals, so the right figure can be confirmed on the page even where the parsed text is wrong.

\paragraph{Writing the tasks.} LLM authoring agents re-derived each candidate figure from the documents and dropped those that did not hold up. For each of the remaining 70, an agent wrote three things. The \emph{question} states the period, units and precision, but not which document to use or how to compute the answer. The \emph{oracle} is a Python program that reads figures from the source documents and computes the key, so no key is typed in by hand. It also checks that the table's totals confirm the key, that every known wrong answer falls outside the range, and, for most tasks, that no document in the archive prints the answer directly. The \emph{known wrong answers} are plausible mistakes: a misread cell, a neighbouring row or year, another base, or a flipped sign. Of the 70 questions, 55 are new and 15 adapt OfficeQA Pro v2 questions \citep{databricks2026officeqaprov2}, asking for a different quantity from the same documents.

\paragraph{Testing the drafts.} A script rejected drafts that lacked an oracle or known wrong answers, had a known wrong answer inside the range, or repeated another question. A Claude Sonnet 5.5 agent then tried each task with the parsed text. It solved 23 and gave a known wrong answer on 39. No task was dropped because of its result.

\subsection{Quality Control}
\label{sec:qc}
\paragraph{Checking the keys on the scans.} Model agents rendered each key's source pages at high resolution, re-read the cells and recomputed the key. All 70 keys held. In 7 tasks the printed figure disagrees with its table's totals; there the key follows the totals.

\paragraph{Revisions.} Before the main run we reworded 8 questions that allowed a second reading and added some of the solver's wrong answers to the lists of known wrong answers. After a pilot, we re-examined the tasks that no attempt had solved. Every key held, and we reworded one question. In three of them the answer could not be read at 60\,ppi in any copy we found, so we replaced their answer pages with sharper images from a public copy, keeping the page numbers and adding no machine-readable text. All other files are unchanged. The authors reviewed every task, every finding of the checking agents and every edit.

\subsection{Grading and Metrics}
\label{sec:grading}
\paragraph{Grading.} A Python grader reads the number on the reply's last \texttt{ANSWER:} line (Appendix~\ref{app:grader}) and passes the attempt if it falls within the task's accepted range. Dollar answers allow about a dollar for rounding; other answers allow a small fraction of a percent. Each range is kept narrow enough to reject the plausible wrong answers, such as a misread digit or the neighbouring year. A reply with no number fails.

\paragraph{Why not an LLM judge.} An LLM judge given the same answer keys passed 32 wrong answers, most of them one digit off. LLM judges can accept wrong answers even when given the reference answer \citep{zheng2023judging}, and near misses are what \bench{} tests.

\paragraph{Outside sources.} Agents can look up benchmark answers online \citep{kapoor2026hal}, so we audited every passing attempt that reached a site outside the archive. Of these 27, 13 used figures from outside the archive and count as failures (Section~\ref{sec:integrity}).

\paragraph{Metrics.} pass@1 is the share of an agent's attempts that are solved; pass@3 is the share of tasks solved at least once in three attempts, and pass$^3$ the share solved all three times \citep{yao2024taubench}. We report 95\% intervals over tasks \citep{miller2024errorbars}, computed by bootstrap. Differences between agents next to each other in the ranking are tested on paired task-level differences \citep{miller2024errorbars}, by bootstrap with a Holm correction.

\paragraph{Access.} Three worked examples, the task in Figure~\ref{fig:example} and two in Appendix~\ref{app:examples}, are given with their answer keys, and the instructions are in Appendix~\ref{app:prompts}. The other tasks stay private to protect their keys.

\section{Experimental Setup}
\label{sec:setup}

We evaluate ten agents, each a model in a command-line harness pinned to one version: Claude Opus 5.5 and Claude Sonnet 5.5 in Claude Code, GPT-6.1 Sol and GPT-5.6 Terra in Codex, Gemini 3.8 Flash and Gemini 4 Argon in Gemini CLI, Grok 4.7 in Grok CLI, and GLM 5.3 Flash, Kimi K3 and the open-weight DeepSeek V4.1 Flash in the open-source Pi harness (versions and providers in Table~\ref{tab:agents}, Appendix~\ref{app:environment}). Reasoning effort is high for every agent. Each agent attempts each task three times, with the same task text, instructions and settings, for 2,100 attempts in all. Cost is the agent's own model spend at list prices, without Reducto's charges (Appendix~\ref{app:effort}). Below we shorten the names to Opus, Sonnet, Sol, Terra, Gemini 3.8 Flash, Argon, Grok, DeepSeek, GLM and Kimi.

\section{Results}
\label{sec:results}

\begin{table}[t]
\centering\small
\caption{\textbf{Main results} on 70 tasks, three attempts per task: pass@1 with its 95\% task-bootstrap interval, pass@3 and pass$^3$ (Section~\ref{sec:grading}); agent-side cost per attempt and in total at list price, with cache reads billed at the cached rate; and median minutes and tool calls per attempt. Best value per column in bold.}
\label{tab:main}
\setlength{\tabcolsep}{3pt}
\begin{tabular}{@{}llccccrrr@{}}
\toprule
Agent & Harness & pass@1 [95\% CI] & pass@3 & pass$^3$ & \$/att. & \$ total & Min. & Calls \\
\midrule
Claude Opus 5.5 & Claude Code & \textbf{64.8} [55.7, 73.3] & \textbf{81.4} & \textbf{45.7} & 1.56 & 328 & 5.3 & 24 \\
Gemini 4 Argon & Gemini CLI & 60.0 [50.0, 70.0] & 71.4 & \textbf{45.7} & 2.45 & 514 & 16.8 & 46 \\
GPT-6.1 Sol & Codex & 51.4 [41.0, 61.4] & 64.3 & 38.6 & 0.62 & 130 & 11.2 & 33 \\
Gemini 3.8 Flash & Gemini CLI & 51.0 [40.5, 61.4] & 65.7 & 37.1 & 1.74 & 366 & 10.3 & 86 \\
Claude Sonnet 5.5 & Claude Code & 41.9 [31.9, 51.9] & 58.6 & 27.1 & 1.19 & 251 & 3.0 & 25 \\
DeepSeek V4.1 Flash & Pi & 37.1 [28.1, 46.2] & 57.1 & 17.1 & 0.25 & 53 & 8.8 & 67 \\
Kimi K3 & Pi & 27.6 [18.6, 36.7] & 40.0 & 18.6 & 2.55 & 535 & 7.6 & 41 \\
GLM 5.3 Flash & Pi & 27.1 [19.0, 35.7] & 45.7 & 11.4 & \textbf{0.24} & \textbf{50} & 12.8 & 60 \\
Grok 4.7 & Grok CLI & 18.1 [11.4, 25.2] & 32.9 & 4.3 & 3.55 & 745 & 7.3 & 47 \\
GPT-5.6 Terra & Codex & 16.7 [10.0, 24.3] & 30.0 & 8.6 & 1.17 & 247 & \textbf{2.4} & \textbf{16} \\
\bottomrule
\end{tabular}
\end{table}

\paragraph{Opus leads, but not significantly.} Opus solves 64.8\% of its attempts and Argon 60.0\% (Table~\ref{tab:main}). The gap is not significant, and four tasks that Opus always solves and Argon never does account for all of it. The agents fall into five pairs: Opus and Argon; Sol and Gemini 3.8 Flash; Sonnet and DeepSeek; Kimi and GLM; Grok and Terra. After correcting for multiple comparisons, no agent is significantly ahead of the one ranked just below it. Two tasks are never solved.

\paragraph{Argon runs out of time; Opus misreads.} When Argon gives a number, it is right 83\% of the time, against 69\% for Opus. But more than a quarter of Argon's attempts hit the 30-minute limit without an answer. Opus is fast. Half of its solves take 5 minutes or less.

\FigHeatmap

\paragraph{Very blurry scans separate the agents.} On clear scans Sol solves the most. On the 20 very blurry tasks, Opus solves 60\% and Argon 50\%, while every other agent solves 37\% or less and Sol drops to 25\% (Figure~\ref{fig:heatmap}). Sol and Argon succeed on different tasks, and together they solve 62 tasks at least once, against 57 for Opus alone.

\paragraph{Agents are consistent on a task, even when wrong.} An agent usually solves a task in all three attempts or in none. Fewer than half as many agent--task pairs have mixed results as chance would predict. Wrong answers repeat too. In 72 pairs, the agent gives the same wrong number every time.

\FigTokens

\paragraph{Cost depends on caching, not accuracy.} GLM is the cheapest per attempt, and GLM, DeepSeek, Sol and Opus give the most accuracy for their cost (Table~\ref{tab:main}). Grok is the most expensive, 14 times DeepSeek's cost per attempt, and solves fewer than half as many attempts. Cost follows uncached input and its price: Argon, Gemini 3.8 Flash and Grok send many uncached tokens, Kimi pays a high rate for its cache reads, and DeepSeek reads almost everything from a cheap cache (Figure~\ref{fig:tokens}). The whole evaluation took about 395 agent-hours and cost \$3,218.

\section{Analysis}
\label{sec:analysis}

\subsection{How Agents Work}
\label{sec:tool-use}

\begin{table}[t]
\centering\small
\caption{\textbf{Tool use} from the agents' traces, which undercount DeepSeek's, Kimi's and GLM's calls and omit Claude Code's sub-agents. \emph{Per attempt}: mean calls and page images viewed. \emph{Reducto}: share of reads that name an uploaded volume by its \texttt{reducto://} identifier (\emph{Upload}), of attempts whose first \texttt{parse\_document} call succeeds (\emph{First read}), and of calls that return an error. \emph{OCR}: attempts in which OCR software run in the shell returned text.}
\label{tab:tooluse}
\setlength{\tabcolsep}{3.5pt}
\begin{tabular}{@{}lrrrrrrrr@{}}
\toprule
 & \multicolumn{4}{c}{Per attempt} & \multicolumn{3}{c}{Reducto (\%)} & \\
\cmidrule(lr){2-5}\cmidrule(lr){6-8}
Agent & Shell & Reducto & Dropbox & Images & Upload & First read & Errors & OCR \\
\midrule
Claude Opus 5.5 & 4.9 & 13.0 & 5.7 & 2.8 & 90.1 & 96.1 & 12.3 & 0 \\
Gemini 4 Argon & 7.5 & 36.8 & 10.9 & 2.2 & 87.4 & 76.2 & 15.1 & 0 \\
GPT-6.1 Sol & 6.6 & 25.0 & 11.5 & 0.0 & 87.3 & 97.1 & 7.5 & 2 \\
Gemini 3.8 Flash & 42.2 & 19.0 & 16.1 & 16.8 & 98.8 & 74.8 & 6.7 & 2 \\
Claude Sonnet 5.5 & 2.5 & 12.2 & 10.0 & 1.1 & 38.3 & 32.0 & 26.6 & 0 \\
DeepSeek V4.1 Flash & 14.6 & 12.0 & 9.5 & 6.4 & 79.8 & 80.1 & 13.6 & 15 \\
Kimi K3 & 14.3 & 18.4 & 11.0 & 4.8 & 95.7 & 36.5 & 14.7 & 59 \\
GLM 5.3 Flash & 17.5 & 21.9 & 12.3 & 6.4 & 78.7 & 15.3 & 36.3 & 28 \\
Grok 4.7 & 21.3 & 12.9 & 15.3 & 8.0 & 82.7 & 89.6 & 18.8 & 35 \\
GPT-5.6 Terra & 7.7 & 15.7 & 9.9 & 0.0 & 53.3 & 20.6 & 23.8 & 3 \\
\bottomrule
\end{tabular}
\end{table}

\paragraph{Agents find files quickly but pass them to Reducto differently.} Every agent typically finds a volume by file name and hands it to the reader within a minute. Opus, Argon, Sol and Gemini 3.8 Flash upload each volume to Reducto once and reuse it; Sonnet and Terra often pass the download link instead and get many Reducto errors. GLM gets the most errors and the fewest successful first reads (Table~\ref{tab:tooluse}).

\paragraph{Answering fast works only for Opus.} Opus, Sonnet, Terra and Grok often answer within 3 minutes, but only Opus is usually right when it does (40 of 56 attempts); Sonnet, Terra and Grok are right in at most 40\% of theirs. Argon never answers that quickly.

\paragraph{Within a harness, the model decides.} Models in the same harness differ widely: by 23 points in Claude Code and 35 in Codex. Opus reuses uploaded volumes in 90\% of its reads and Sonnet in 38\%. Gemini 3.8 Flash looks at page images in most of its attempts and Argon in under a third, and neither Codex agent ever does. Only Claude Code starts sub-agents, and 16 of its attempts end with a note that they are still working, which the instructions forbid.

\paragraph{Looking at the page helps less than it seems.} Attempts that look at a page image are solved more often (42\% against 29\%), but among attempts that run 5 minutes or more the gap shrinks and is no longer significant. Gemini 3.8 Flash looks at the page in nearly all of its very blurry attempts and still misreads more than half of the answers it gives there.

\paragraph{Agents trust the printed figure over the totals.} On the 7 tasks whose printed figure is wrong, agents solve 24\% of attempts, against 41\% elsewhere, and most failures there report the printed figure. In 30 failed replies the agent's own check pointed to the right answer, and it still reported the printed figure. Opus once wrote down the correct total and then answered: ``My answer uses the printed figure.''

\subsection{Failed Attempts}
\label{sec:failures}

\FigFailures

\paragraph{Most failures are wrong numbers, often one digit off.} About three quarters of failed attempts give a wrong number and the rest mostly give none. More than a quarter of the wrong numbers differ from the key in a single digit. A model agent assigned each failure one cause, and an independent re-check agreed on 93\% of a sample (Appendix~\ref{app:failures}). Misread digits are the most common cause, 36\% of failures, and lead for seven of the ten agents (Figure~\ref{fig:failures}). They affect 7\% of answers on clear scans and 56\% on very blurry ones, most often an 8 confused with a 6 or a 9. Next come running out of time and reporting a printed figure that the totals contradict.

\paragraph{Agents make the same mistakes.} Most wrong numbers (62\%) are also given by another agent. On 20 tasks the most common answer is wrong, so a majority vote would solve at most 50 tasks, while at least one attempt solves 68. Most of the common ones were not on our lists: of the 38 wrong numbers given by four or more agents, 29 are not among the task's known wrong answers. Only 20 of the 303 known wrong answers were ever given.

\paragraph{Grok answers without reading.} Grok gave a number before reading any page in 36 attempts, often right after a failed upload, and these answers are mostly far off. In about half of its failures, the figures in its reply appear in nothing it read. One reply cites a file that does not exist; another answers 1234.56.

\subsection{Integrity}
\label{sec:integrity}
Following the Agentic Benchmark Checklist \citep{zhu2025abc} (Appendix~\ref{app:checklist}), we checked whether an agent could score without doing the task. Agents' shells could reach the web, and 13 passing attempts took their figures from another copy of a volume, a statute or the benchmark's own materials. These count as failures, and limiting the sites agents can reach would prevent them. In 4 of Grok's solves the figures appear in no tool output; they count, because the grader checks only the number. No agent changed an archive file.

\section{Limitations}
\label{sec:limitations}

\paragraph{Tasks and keys.} The tasks are not a random sample of questions: each is built on a table cell that the page's totals can confirm, which favours faint, hard-to-read figures, and all come from one English-language archive with a single number as the answer. The keys were checked by model agents, mostly Claude Opus 5.5, the top-scoring model, rather than by domain experts. Claude Sonnet 5.5 tried the drafts and is also one of the evaluated agents. Agents solve fewer of the tasks on which it gave a known wrong answer. One question allows a second reading that our key marks wrong; accepting it would change no rank.

\paragraph{Grading and setup.} The grader checks only the number: a dollar answer within \$1.00 of the key passes even with a misread cents digit, and an abstention scores the same as a wrong figure. A score measures one model in one harness reading through one hosted reader, Reducto, whose version is not pinned. Another harness could change scores and their order \citep{li2026modelharness}, and so could another reader. A longer time limit could help Argon and GLM, which often ran out of time, though Argon's own work held the answer in only 7 of its 58 time-outs.

\paragraph{Size and exposure.} With 70 tasks, pass@1 intervals are wide, 14 to 21 points. The PDFs and the OfficeQA Pro v2 questions are available to anyone who requests access \citep{databricks2026officeqaprov2}, and frontier models have memorised many historical financial statistics \citep{opsahlong2026officeqapro}. Keys that no document prints limit, but do not rule out, answering from memory, and agents solve adapted and new questions about equally often. We ran no human baseline.

\section{Conclusion}
\label{sec:conclusion}

\bench{} asks an agent to find, read and report one figure from scanned Treasury records through a drive's MCP tools and a hosted OCR reader. The two leading agents, Claude Opus 5.5 and Gemini 4 Argon, solve 64.8\% and 60.0\% of their attempts: Argon is usually right when it answers but often runs out of time, while Opus answers fast and misreads more. Most failures are wrong numbers, often one digit off, and agents tend to report the printed figure even when their own check contradicts it. Because a script grades every reply, these results do not depend on a model's judgement.

\section*{Disclosure}
The authors work at Veris AI, which builds the Dropbox twin used here. \bench{} is released together with Reducto, whose hosted reader every agent used.

\section*{Reproducibility Statement}
Each agent is a fixed model identifier, provider, harness version and reasoning effort (Table~\ref{tab:agents}), and every attempt ran the same task text and agent configuration on a fresh copy of the twin. The instructions are given verbatim in Appendix~\ref{app:prompts}. Reducto's hosted service reports no release version. Its r-1 model made the parses, and all calls fell between 3 and 21 October 2026. Grading every stored reply again reproduces all 2,100 verdicts, and costs use October 2026 list prices (Appendix~\ref{app:effort}). The instructions (Appendix~\ref{app:prompts}) and three worked examples with their answer keys (Figure~\ref{fig:example}, Appendix~\ref{app:examples}) are given in this paper.

\section*{LLM Usage Statement}
LLM agents drafted the tasks, their answer-key programs and known wrong answers, checked every key on the scans and labelled failure causes (Sections~\ref{sec:construction} and~\ref{sec:qc}). The authors reviewed every task, finding and edit. The authors used LLM assistants to write analysis scripts and to edit this paper; scripts computed every number from the evaluation records.

\label{endofmaintext}

{\footnotesize
\setlength{\bibsep}{1.5pt plus 0.5pt}
\bibliographystyle{plainnat}
\bibliography{references}

\begin{thebibliography}{26}
\providecommand{\natexlab}[1]{#1}
\providecommand{\url}[1]{\texttt{#1}}
\expandafter\ifx\csname urlstyle\endcsname\relax
  \providecommand{\doi}[1]{doi: #1}\else
  \providecommand{\doi}{doi: \begingroup \urlstyle{rm}\Url}\fi

\bibitem[{Anthropic}(2024)]{anthropic2024mcp}
{Anthropic}.
\newblock Introducing the {Model Context Protocol}.
\newblock \url{https://www.anthropic.com/news/model-context-protocol}, 2024.
\newblock Published 2024-11-25. Accessed 2026-10-12.

\bibitem[Barres et~al.(2026)Barres, Dong, Ray, Si, and
  Narasimhan]{barres2025tau2}
Victor Barres, Honghua Dong, Soham Ray, Xujie Si, and Karthik Narasimhan.
\newblock {$\tau^2$-Bench}: Evaluating conversational agents in a dual-control
  environment.
\newblock In \emph{Proceedings of the 43rd International Conference on Machine
  Learning (ICML)}, volume 306, pages 6736--6782, 2026.

\bibitem[Bigeard et~al.(2025)Bigeard, Nashold, Krishnan, and
  Wu]{bigeard2025financeagent}
Antoine Bigeard, Langston Nashold, Rayan Krishnan, and Shirley Wu.
\newblock Finance agent benchmark: Benchmarking {LLMs} on real-world financial
  research tasks.
\newblock \emph{arXiv preprint arXiv:2508.00828}, 2025.

\bibitem[Chen et~al.(2021)Chen, Chen, Smiley, Shah, Borova, Langdon, Moussa,
  Beane, Huang, Routledge, and Wang]{chen2021finqa}
Zhiyu Chen, Wenhu Chen, Charese Smiley, Sameena Shah, Iana Borova, Dylan
  Langdon, Reema Moussa, Matt Beane, Ting-Hao Huang, Bryan Routledge, and
  William~Yang Wang.
\newblock {FinQA}: A dataset of numerical reasoning over financial data.
\newblock In \emph{Proceedings of the 2021 Conference on Empirical Methods in
  Natural Language Processing (EMNLP)}, pages 3697--3711, 2021.
\newblock \doi{10.18653/v1/2021.emnlp-main.300}.

\bibitem[{Databricks}(2026)]{databricks2026officeqaprov2}
{Databricks}.
\newblock {OfficeQA Pro v2}: A grounded reasoning benchmark.
\newblock \url{https://huggingface.co/datasets/databricks/officeqa-pro-v2},
  2026.
\newblock Hugging Face dataset; data CC BY-SA 4.0, code Apache 2.0; gated
  access. Accessed 2026-10-12.

\bibitem[{Dropbox}(2026)]{dropbox2026mcp}
{Dropbox}.
\newblock How to connect the {Dropbox} remote {MCP} server.
\newblock
  \url{https://help.dropbox.com/integrations/connect-dropbox-mcp-server}, 2026.
\newblock Dropbox Help Center, page updated 2026-09-30. Accessed 2026-10-12.

\bibitem[Guo et~al.(2026)Guo, Zhang, Zhang, Li, Zheng, Lei, Peng, Xi, Gui, and
  Zhang]{guo2026agora}
Honglin Guo, Qi~Zhang, Yu~Zhang, Weijie Li, Rui Zheng, Zhikai Lei, Qiyuan Peng,
  Zhiheng Xi, Tao Gui, and Qi~Zhang.
\newblock {AGORA}: An archive-grounded benchmark for agentic workplace document
  reasoning.
\newblock \emph{arXiv preprint arXiv:2606.24526}, 2026.

\bibitem[Islam et~al.(2023)Islam, Kannappan, Kiela, Qian, Scherrer, and
  Vidgen]{islam2023financebench}
Pranab Islam, Anand Kannappan, Douwe Kiela, Rebecca Qian, Nino Scherrer, and
  Bertie Vidgen.
\newblock {FinanceBench}: A new benchmark for financial question answering.
\newblock \emph{arXiv preprint arXiv:2311.11944}, 2023.

\bibitem[Kapoor et~al.(2026)Kapoor, Stroebl, Kirgis, Nadgir, Siegel,
  et~al.]{kapoor2026hal}
Sayash Kapoor, Benedikt Stroebl, Peter Kirgis, Nitya Nadgir, Zachary~S. Siegel,
  et~al.
\newblock Holistic agent leaderboard: The missing infrastructure for {AI} agent
  evaluation.
\newblock In \emph{International Conference on Learning Representations
  (ICLR)}, 2026.

\bibitem[Li et~al.(2026)Li, Zhou, Teng, Yang, Deng, Lyu, Dong, Chen, and
  An]{li2026modelharness}
Yixuan Li, Yiyun Zhou, Yao~Long Teng, Fuchao Yang, Yanchen Deng, Zhiyi Lyu,
  Xuyu Dong, Feng Chen, and Bo~An.
\newblock Finding the right fit: Model-harness interactions across agent tasks.
\newblock \emph{arXiv preprint arXiv:2610.00917}, 2026.

\bibitem[Ma et~al.(2024)Ma, Zang, Chen, Chen, Jiao, Li, Lu, Liu, Ma, Dong,
  Zhang, Pan, Jiang, Wang, Cao, and Sun]{ma2024mmlongbenchdoc}
Yubo Ma, Yuhang Zang, Liangyu Chen, Meiqi Chen, Yizhu Jiao, Xinze Li, Xinyuan
  Lu, Ziyu Liu, Yan Ma, Xiaoyi Dong, Pan Zhang, Liangming Pan, Yu-Gang Jiang,
  Jiaqi Wang, Yixin Cao, and Aixin Sun.
\newblock {MMLongBench-Doc}: Benchmarking long-context document understanding
  with visualizations.
\newblock In \emph{Advances in Neural Information Processing Systems (Datasets
  and Benchmarks Track)}, volume~37, pages 95963--96010, 2024.

\bibitem[Mathew et~al.(2021)Mathew, Karatzas, and Jawahar]{mathew2021docvqa}
Minesh Mathew, Dimosthenis Karatzas, and C.~V. Jawahar.
\newblock {DocVQA}: A dataset for {VQA} on document images.
\newblock In \emph{Proceedings of the IEEE/CVF Winter Conference on
  Applications of Computer Vision (WACV)}, pages 2200--2209, 2021.

\bibitem[Merrill et~al.(2026)Merrill, Shaw, Carlini, Li, Raj,
  et~al.]{merrill2026terminalbench}
Mike~A. Merrill, Alexander~G. Shaw, Nicholas Carlini, Boxuan Li, Harsh Raj,
  et~al.
\newblock {Terminal-Bench}: Benchmarking agents on hard, realistic tasks in
  command line interfaces.
\newblock In \emph{International Conference on Learning Representations
  (ICLR)}, 2026.

\bibitem[Mialon et~al.(2024)Mialon, Fourrier, Swift, Wolf, LeCun, and
  Scialom]{mialon2024gaia}
Gr{\'e}goire Mialon, Cl{\'e}mentine Fourrier, Craig Swift, Thomas Wolf, Yann
  LeCun, and Thomas Scialom.
\newblock {GAIA}: a benchmark for general {AI} assistants.
\newblock In \emph{International Conference on Learning Representations
  (ICLR)}, 2024.

\bibitem[Miller(2024)]{miller2024errorbars}
Evan Miller.
\newblock Adding error bars to evals: A statistical approach to language model
  evaluations.
\newblock \emph{arXiv preprint arXiv:2411.00640}, 2024.

\bibitem[Opsahl-Ong et~al.(2026)Opsahl-Ong, Singhvi, Collins, Zhou, Wang,
  Baheti, Oertell, Portes, Havens, Elsen, Bendersky, Zaharia, and
  Chen]{opsahlong2026officeqapro}
Krista Opsahl-Ong, Arnav Singhvi, Jasmine Collins, Ivan Zhou, Cindy Wang,
  Ashutosh Baheti, Owen Oertell, Jacob Portes, Sam Havens, Erich Elsen, Michael
  Bendersky, Matei Zaharia, and Xing Chen.
\newblock {OfficeQA Pro}: An enterprise benchmark for end-to-end grounded
  reasoning.
\newblock \emph{arXiv preprint arXiv:2603.08655}, 2026.

\bibitem[{Reducto}(2026)]{reducto2026mcp}
{Reducto}.
\newblock Reducto {MCP} server.
\newblock \url{https://docs.reducto.ai/mcp-server}, 2026.
\newblock Accessed 2026-10-12.

\bibitem[Stanis{\l}awek et~al.(2021)Stanis{\l}awek, Grali{\'n}ski,
  Wr{\'o}blewska, Lipi{\'n}ski, Kaliska, Rosalska, Topolski, and
  Biecek]{stanislawek2021kleister}
Tomasz Stanis{\l}awek, Filip Grali{\'n}ski, Anna Wr{\'o}blewska, Dawid
  Lipi{\'n}ski, Agnieszka Kaliska, Paulina Rosalska, Bartosz Topolski, and
  Przemys{\l}aw Biecek.
\newblock Kleister: Key information extraction datasets involving long
  documents with complex layouts.
\newblock In \emph{Document Analysis and Recognition -- ICDAR 2021}, Lecture
  Notes in Computer Science, pages 564--579, 2021.
\newblock \doi{10.1007/978-3-030-86549-8_36}.

\bibitem[Wei et~al.(2025)Wei, Sun, Papay, McKinney, Han, Fulford, Chung,
  Passos, Fedus, and Glaese]{wei2025browsecomp}
Jason Wei, Zhiqing Sun, Spencer Papay, Scott McKinney, Jeffrey Han, Isa
  Fulford, Hyung~Won Chung, Alex~Tachard Passos, William Fedus, and Amelia
  Glaese.
\newblock {BrowseComp}: A simple yet challenging benchmark for browsing agents.
\newblock \emph{arXiv preprint arXiv:2504.12516}, 2025.

\bibitem[Wu et~al.(2026)Wu, Liu, Zhang, Chen, Meng, et~al.]{wu2025mcpmark}
Zijian Wu, Xiangyan Liu, Xinyuan Zhang, Lingjun Chen, Fanqing Meng, et~al.
\newblock {MCPMark}: A benchmark for stress-testing realistic and comprehensive
  {MCP} use.
\newblock In \emph{International Conference on Learning Representations
  (ICLR)}, 2026.

\bibitem[Xu et~al.(2025)Xu, Song, Li, Tang, Jain,
  et~al.]{xu2025theagentcompany}
Frank~F. Xu, Yufan Song, Boxuan Li, Yuxuan Tang, Kritanjali Jain, et~al.
\newblock {TheAgentCompany}: Benchmarking {LLM} agents on consequential real
  world tasks.
\newblock In \emph{Advances in Neural Information Processing Systems (Datasets
  and Benchmarks Track)}, 2025.

\bibitem[Yao et~al.(2025)Yao, Shinn, Razavi, and Narasimhan]{yao2024taubench}
Shunyu Yao, Noah Shinn, Pedram Razavi, and Karthik Narasimhan.
\newblock {$\tau$-bench}: A benchmark for {Tool-Agent-User} interaction in
  real-world domains.
\newblock In \emph{International Conference on Learning Representations
  (ICLR)}, 2025.

\bibitem[Zhang et~al.(2025)Zhang, Zhang, Wang, Ouyang, Wen, Li, Chow, He, and
  Zhang]{zhang2025ohrbench}
Junyuan Zhang, Qintong Zhang, Bin Wang, Linke Ouyang, Zichen Wen, Ying Li,
  Ka-Ho Chow, Conghui He, and Wentao Zhang.
\newblock {OCR} hinders {RAG}: Evaluating the cascading impact of {OCR} on
  retrieval-augmented generation.
\newblock In \emph{Proceedings of the IEEE/CVF International Conference on
  Computer Vision (ICCV)}, pages 17443--17453, 2025.
\newblock \doi{10.1109/ICCV51701.2025.01620}.

\bibitem[Zheng et~al.(2023)Zheng, Chiang, Sheng, Zhuang, Wu, Zhuang, Lin, Li,
  Li, Xing, Zhang, Gonzalez, and Stoica]{zheng2023judging}
Lianmin Zheng, Wei-Lin Chiang, Ying Sheng, Siyuan Zhuang, Zhanghao Wu, Yonghao
  Zhuang, Zi~Lin, Zhuohan Li, Dacheng Li, Eric~P. Xing, Hao Zhang, Joseph~E.
  Gonzalez, and Ion Stoica.
\newblock Judging {LLM}-as-a-judge with {MT-Bench} and {Chatbot Arena}.
\newblock In \emph{Advances in Neural Information Processing Systems (Datasets
  and Benchmarks Track)}, volume~36, pages 46595--46623, 2023.

\bibitem[Zhu et~al.(2022)Zhu, Lei, Feng, Wang, Zhang, and Chua]{zhu2022tatdqa}
Fengbin Zhu, Wenqiang Lei, Fuli Feng, Chao Wang, Haozhou Zhang, and Tat-Seng
  Chua.
\newblock Towards complex document understanding by discrete reasoning.
\newblock In \emph{Proceedings of the 30th ACM International Conference on
  Multimedia (ACM MM)}, pages 4857--4866, 2022.
\newblock \doi{10.1145/3503161.3548422}.

\bibitem[Zhu et~al.(2025)Zhu, Jin, Pruksachatkun, Zhang, Liu, Cui, Kapoor,
  Longpre, Meng, Weiss, Barez, Gupta, Dhamala, Merizian, Giulianelli, Coppock,
  Ududec, Kellermann, Sekhon, Steinhardt, Schwettmann, Narayanan, Zaharia,
  Stoica, Liang, and Kang]{zhu2025abc}
Yuxuan Zhu, Tengjun Jin, Yada Pruksachatkun, Andy~K. Zhang, Shu Liu, Sasha Cui,
  Sayash Kapoor, Shayne Longpre, Kevin Meng, Rebecca Weiss, Fazl Barez, Rahul
  Gupta, Jwala Dhamala, Jacob Merizian, Mario Giulianelli, Harry Coppock,
  Cozmin Ududec, Antony Kellermann, Jasjeet~S. Sekhon, Jacob Steinhardt, Sarah
  Schwettmann, Arvind Narayanan, Matei Zaharia, Ion Stoica, Percy Liang, and
  Daniel Kang.
\newblock Establishing best practices in building rigorous agentic benchmarks.
\newblock In \emph{Advances in Neural Information Processing Systems (Datasets
  and Benchmarks Track)}, volume~38, 2025.
\newblock \doi{10.52202/085713-5547}.

\end{thebibliography}
}

\appendix
\raggedbottom   %
\providecommand{\FloatBarrier}{}

\section{Instructions}
\label{app:prompts}

The shared instructions follow verbatim (1,402 characters, the same for all ten agents). The first two sentences of the second paragraph are OfficeQA Pro's \citep{opsahlong2026officeqapro}.

\begin{promptbox}[Instructions]
\raggedright
You answer questions about the historical U.S. Treasury document corpus stored in the company Dropbox. The files are scanned PDFs; their filenames carry the document family and year, for example combined\_statement\_\_historical\_\_cs-1941.pdf, govinfo\_receipts\_\_1832\_\_GOVPUB-....pdf, combined\_statement\_\_modern\_\_2016\_\_c07.pdf. Read the scanned pages with the Reducto tools. Volumes run to 1,000 pages, so locate the right pages first, for example via the contents or index pages, and read in small page ranges. To read a page range with parse\_document, pass options with settings.page\_range set to an object holding the start and end page numbers, for example start 43 and end 47. Ground every figure you use in what the documents contain.\par
\smallskip
If multiple documents report the same metric, use the most up-to-date revision unless the question specifies an exact date or document. Do not rely on the first matching value you encounter. Use maximum precision in your calculations unless the question says otherwise, and give the numeric answer at exactly the precision the question requests. If the corpus does not contain the answer, say so. Reply with exactly one complete message per question: never ask the user to wait, promise to report back later, or offer to send the document itself. Name the file(s) you used, then end your message with a single final line in exactly this form:\par
ANSWER: <final answer>
\end{promptbox}

\section{Worked Examples}
\label{app:examples}
Besides the task in Figure~\ref{fig:example}, two tasks are public worked examples.

\begin{examplebox}[Worked example P004]
\textbf{Question.} ``By how many dollars did the appropriations made for the Department of State's Foreign Intercourse accounts for their own fiscal year change from fiscal year 1923 to fiscal year 1924? Count only appropriations chargeable to the fiscal year in question, excluding continuing (no-year) and prior-year appropriations and excluding balances brought forward from earlier years. Report the change as the fiscal 1924 figure minus the fiscal 1923 figure, in dollars to the cent. Include appropriations covering two fiscal years (for example, 1923-24) in the fiscal year in which they begin.''

\smallskip
\textbf{Where.} \nolinkurl{combined_statement__historical__cs-1923.pdf} and \nolinkurl{cs-1924.pdf}, the Department of State's Foreign Intercourse accounts (printed pages 149--154 and 148--154), Appropriations column.

\smallskip
\textbf{Solution.} Summing the appropriations chargeable to each year, with two-year appropriations counted in their first year and no-year and prior-year appropriations and balances brought forward left out, gives \$8,768,393.16 for fiscal 1923 and \$8,741,552.24 for fiscal 1924, a change of $-$\$26,840.92.

\smallskip
\textbf{Key.} $-$\$26,840.92 $\pm$ \$1.00. \textbf{Known wrong answers} (4) include 26,840.92 (the sign flipped) and $-$21,840.92 (one digit misread).

\smallskip
\textbf{Outcome.} Solved in 15 of 30 attempts; 7 answered 63,159.08, after summing fiscal 1923 to \$8,678,393.16, \$90,000 short.
\end{examplebox}

\begin{examplebox}[Worked example P050]
\textbf{Question.} ``By what percentage did net outlays for the Bureau of the Census salaries and expenses account change from fiscal year 1979 to fiscal year 1980? Express the change as a percentage of the fiscal year 1979 amount, rounded to two decimal places.''

\smallskip
\textbf{Where.} \nolinkurl{combined_statement__historical__cs-1979.pdf} (printed page 194) and \nolinkurl{cs-1980.pdf} (printed page 190), the Bureau of the Census salaries and expenses account, its net outlays subtotal.

\smallskip
\textbf{Solution.} The fiscal 1979 subtotal is printed as \$50,521,194.74, but the account's own rows, and its appropriation less the amount withdrawn, both give \$50,531,194.75. With \$55,134,857.17 for fiscal 1980, the change is $(55{,}134{,}857.17 - 50{,}531{,}194.75) / 50{,}531{,}194.75 = 9.11\%$.

\smallskip
\textbf{Key.} 9.11\% (9.1054 to 9.1146 accepted). \textbf{Known wrong answers} (4) include 9.13, from the printed fiscal 1979 subtotal.

\smallskip
\textbf{Outcome.} Solved in 17 of 30 attempts; 8 answered 9.13, using the printed subtotal.
\end{examplebox}

\FloatBarrier
\section{Construction and Answer Keys}
\label{app:construction}

Oracle-first construction (Section~\ref{sec:construction}) began after an audit of seven OfficeQA Pro questions found one whose key the parsed text made unreachable. The 70 oracles hold 11,339 lines and 1,078 assertions. After the run, eleven dollar ranges were narrowed, which changed no verdict.

Table~\ref{tab:buckets} sorts the 70 tasks into seven buckets by what the question asks the agent to do. Pass@1 is lowest, 25.6\%, when the question asks for the difference of two items in one year, and highest, 63.3\%, when it asks for a percentage change or excess.

\begin{table}[H]
\centering
\footnotesize
\caption{Tasks by what the question asks the agent to do. Very blurry and multi-document are shares of the bucket's tasks. Of the 70 tasks, 60 list a misprinted or misread figure among their known wrong answers. The trap column names the most common other kind of known wrong answer in the bucket, with the number of its tasks that carry it, or the misprinted or misread figure when no other kind reaches a quarter of the bucket. Pass@1 is over all attempts by the ten agents on the bucket's tasks; the best agent has the highest pass@1 on them.}
\label{tab:buckets}
\setlength{\tabcolsep}{2.5pt}
\begin{tabular*}{\linewidth}{@{\extracolsep{\fill}}lrrrlrl@{}}
\toprule
What the question asks & Tasks & \begin{tabular}[b]{@{}r@{}}Very\\blurry\end{tabular} & \begin{tabular}[b]{@{}r@{}}Multi-\\document\end{tabular} & Typical trap & Pass@1 & Best agent \\
\midrule
Read one printed figure & 7 & 43\% & 0\% & Neighbouring row or total (3) & 37.1\% & Opus (76.2\%) \\
Add up lines of one table & 10 & 40\% & 0\% & Misprinted or misread figure (10) & 36.3\% & Argon (86.7\%) \\
Difference of two items in one year & 12 & 42\% & 8\% & Sign or direction (6) & 25.6\% & Argon (38.9\%) \\
Change between two periods & 12 & 33\% & 75\% & Sign or direction (8) & 40.3\% & Opus (77.8\%) \\
Share of a total & 13 & 23\% & 8\% & Wrong base (9) & 44.1\% & Opus (69.2\%) \\
Percentage change or excess & 5 & 20\% & 60\% & Wrong base (2) & 63.3\% & Sonnet (86.7\%) \\
Statistic or pick over many years & 11 & 0\% & 54\% & Another edition's figure (3) & 42.4\% & Sol (78.8\%) \\
\midrule
All tasks & 70 & 29\% & 29\% & & 39.6\% & \\
\bottomrule
\end{tabular*}
\end{table}

\paragraph{How keys are computed.} Each oracle works at full precision, sums dollar amounts in exact cents and rounds once, at the end, to the precision the question asks for. Lookups are also recomputed from the table's totals, and where a series is reprinted in later volumes the oracle reads every printing among its sources. A second range is accepted only for a second reasonable reading of the question, and a figure that fails its own table's printed totals is never accepted.

\FloatBarrier
\section{Environment}
\label{app:environment}

\begin{table}[H]
\centering\footnotesize
\caption{\textbf{Agents.} Each agent is a model, called by the identifier shown through the provider shown, in a command-line harness pinned to one version; effort is the reasoning effort set through the harness. Codex runs with its own sandbox off inside the container.}
\label{tab:agents}
\setlength{\tabcolsep}{3.5pt}
\begin{tabular}{@{}lllll@{}}
\toprule
Agent & Model identifier & Harness & Provider & Effort \\
\midrule
Claude Opus 5.5 & \texttt{us.anthropic.claude-opus-5-5} & Claude Code 2.1.284 & Amazon Bedrock & high \\
Claude Sonnet 5.5 & \texttt{us.anthropic.claude-sonnet-5-5} & Claude Code 2.1.284 & Amazon Bedrock & high \\
GPT-6.1 Sol & \texttt{openai.gpt-6.1-sol} & Codex 0.153.3 & Amazon Bedrock & high \\
GPT-5.6 Terra & \texttt{openai.gpt-5.6-terra} & Codex 0.153.3 & Amazon Bedrock & high \\
Gemini 3.8 Flash & \texttt{gemini-3.8-flash} & Gemini CLI 0.61.0 & Gemini API & high \\
Gemini 4 Argon & \texttt{gemini-4-argon} & Gemini CLI 0.61.0 & Gemini API & high \\
Grok 4.7 & \texttt{global.xai.grok-4.7} & Grok CLI 1.0.46 & Amazon Bedrock & high \\
DeepSeek V4.1 Flash & \texttt{deepseek-ai/DeepSeek-V4.1-Flash} & Pi 0.99.2 & Baseten & high \\
GLM 5.3 Flash & \texttt{accounts/fireworks/models/glm-5p3-flash} & Pi 0.99.2 & Fireworks & high \\
Kimi K3 & \texttt{moonshotai/Kimi-K3} & Pi 0.99.2 & Baseten & high \\
\bottomrule
\end{tabular}
\end{table}

\paragraph{Archive and container.} The 1,435 files hold 118,524 pages (13.3\,GB): 1,321 \emph{Combined Statements} (1872--2024) and 114 earlier reports (1793--1893), of up to 1,016 pages. Of these, 1,256 are scanned at 150\,ppi, 102 at 100--120 and 77 at 60--80; 582 exceed the 5\,MB limit of \texttt{get\_file\_content}. Links from \texttt{download\_link} last 1 to 15 minutes and, per Dropbox's tool description, work once. Each attempt runs in an unprivileged Debian~12 container with 2 CPUs and 4\,GiB, which provides Python, poppler, ripgrep, \texttt{curl} and \texttt{jq}. The harness stops the CLI a minute before the 30-minute limit so that its reply is still recorded. Reducto parses with its r-1 model, and seven of its tools were available.

\paragraph{Construction and statistics.} The candidate search scanned the 206,820 tables in OfficeQA Pro v2's parsed text, catalogued 201 cells confirmed by their table's totals and tied 36 of the 90 OfficeQA Pro v2 questions to such a cell. The authoring agents received 110 candidates (90 catalogued cells and 20 OfficeQA Pro v2 questions) and dropped 40. The tasks draw on 55 volumes with a median of 594 pages. Bootstrap intervals use 4,000 percentile resamples over tasks, keeping each task's three attempts together; the Holm correction is over the nine comparisons between neighbours in the ranking.

Two of Reducto's tool descriptions mislead: \texttt{parse\_document} offers an \texttt{agentic} option that the server accepts only with a prompt (1,317 refused calls), and \texttt{upload\_file} says it accepts local paths, which the hosted server cannot read (411 refused calls, 225 of them Grok's and 123 GLM's).

\FloatBarrier
\section{Grader}
\label{app:grader}

\paragraph{Extraction rule.} The grader strips markdown and LaTeX wrappers, reads Unicode minus signs, a dash attached to a digit or currency sign, and accounting parentheses as negative, and never rescales a number by a unit word. It uses the last line that starts with \texttt{ANSWER} or \texttt{Final answer}, in any case. On it, it drops numbers that are part of a file name, follow a word such as page, table or line, or are a year after a word such as fiscal year or June 30, and takes the first remaining number, or the last if the line states a correction (``should be'', ``instead'', an arrow). Abstention wording before the number (``not available'', ``cannot be determined'') makes the reply abstain. Without an answer line, it takes the last amount: a number with a currency sign, decimal point, thousands separator or percent sign that is not a year. If the final reply was not saved, it reads the agent's last message from the trace.

\FloatBarrier
\section{Failure Causes}
\label{app:failures}

A model agent assigned each failed attempt one primary cause from its full record (Section~\ref{sec:failures}). \emph{Digits}: a figure misread from the scan or from Reducto's text and not checked. \emph{Kept}: a figure reported as printed although the table's own totals point to the key. \emph{Source}: the wrong volume, an earlier edition whose figures were later restated, or figures from outside the archive. \emph{Row}: the right table but the wrong line, column or period. \emph{Method}: the right inputs but the wrong computation or definition. \emph{Ungrounded}: the reply's figures appear in no tool output of the attempt. \emph{Gave up}: the agent declined or said the corpus lacks the data. \emph{Time}: a time limit ended the attempt without a committed number. \emph{Ended}: the attempt ended earlier without a committed number.

\FloatBarrier
\section{Cost and Effort}
\label{app:effort}

Table~\ref{tab:app-effort} gives each agent's time, tokens and total cost. Prices per million tokens of input, cache read, cache write and output are the standard text-token list prices, with short context and five-minute cache writes: Opus \$4, 0.20, 5, 20; Sonnet \$2, 0.20, 2.50, 10; Sol \$2, 0.10, 2.50, 10; Terra \$2, 0.20, 2.50, 12; Gemini 3.8 Flash \$0.75, 0.075, 0.75, 3.75; Argon \$2, 0.10, none, 10; Grok \$2, 0.50, 2, 6; DeepSeek \$0.30, 0.007, 0.30, 1.20; Kimi \$3, 0.30, 3, 15; and GLM \$0.15, 0.03, 0.15, 0.50, from the 18 October list, since the list used for the others has no GLM price. Regional-endpoint uplifts and long-context surcharges are not applied. One Kimi attempt recorded no tokens or cost; it counts as zero in the totals and means.

\begin{table}[H]
\centering\footnotesize
\setlength{\tabcolsep}{4pt}
\caption{\textbf{Cost and tokens per agent.} \emph{Hours}: total agent time; tokens and mean cost are per attempt, with uncached input, cache reads and cache writes making up the input, and reasoning counted within output; cost is agent-side at list price; \emph{All} gives means over all attempts.}
\label{tab:app-effort}
\begin{tabular}{@{}lrrrrrrrr@{}}
\toprule
 & & \multicolumn{3}{c}{Input} & \multicolumn{2}{c}{Output (k)} & \multicolumn{2}{c}{Cost (\$)} \\
\cmidrule(lr){3-5}\cmidrule(lr){6-7}\cmidrule(l){8-9}
Agent & Hours & Uncached (k) & Cache read (M) & Cache write (M) & All & Reasoning & Mean & Total \\
\midrule
Claude Opus 5.5 & 24.9 & 3.0 & 2.31 & 0.15 & 16.1 & 5.1 & 1.56 & 327.91 \\
Gemini 4 Argon & 64.7 & 910.6 & 3.32 & 0.00 & 34.4 & 24.3 & 2.45 & 514.01 \\
GPT-6.1 Sol & 44.4 & 4.1 & 2.69 & 0.11 & 6.0 & 2.0 & 0.62 & 129.51 \\
Gemini 3.8 Flash & 42.5 & 867.7 & 12.57 & 0.00 & 46.6 & 33.2 & 1.74 & 366.11 \\
Claude Sonnet 5.5 & 17.5 & 0.1 & 2.81 & 0.18 & 17.1 & 5.4 & 1.19 & 250.51 \\
DeepSeek V4.1 Flash & 45.6 & 124.9 & 13.61 & 0.00 & 100.8 & 88.7 & 0.25 & 53.28 \\
Kimi K3 & 39.1 & 267.9 & 5.00 & 0.00 & 16.4 & 7.9 & 2.55 & 535.25 \\
GLM 5.3 Flash & 53.9 & 233.2 & 6.30 & 0.00 & 26.1 & 19.7 & 0.24 & 49.77 \\
Grok 4.7 & 41.7 & 970.5 & 2.92 & 0.00 & 23.9 & 15.7 & 3.55 & 744.79 \\
GPT-5.6 Terra & 21.1 & 11.1 & 3.19 & 0.17 & 8.0 & 4.7 & 1.17 & 246.57 \\
\midrule
All & 395.4 & 339.3 & 5.47 & 0.06 & 29.5 & 20.7 & 1.53 & 3,217.71 \\
\bottomrule
\end{tabular}
\end{table}

\FloatBarrier
\section{Agentic Benchmark Checklist}
\label{app:checklist}

\bench{} against the Agentic Benchmark Checklist \citep{zhu2025abc}; of the outcome-validity items, those for answer matching (plus O.b.2) apply. Status: Yes, Partly or No.

\begin{center}\scriptsize
\setlength{\tabcolsep}{3pt}
\begin{tabular}{@{}llll@{}}
\toprule
Item & Requirement & Status & Evidence or gap \\
\midrule
T.1--2 & Tool versions; API tools accessible & Partly & CLI versions, model ids and image stated; Reducto (r-1) is hosted, not pinned. \\
T.3--4 & API failures handled; state cleared & Yes & 30-minute limit; every attempt scored; fresh twin and container. \\
T.5 & Isolated from ground truth & Partly & Key never in the container; outside-source passes fail. \\
T.6 & Setup frozen over time & No & Reducto is live; its tool list changed in the run. \\
T.7--8 & Ground truth verified; solvable & Yes & Oracle-computed keys, confirmed on scans; 68 tasks solved. \\
T.9--10 & Oracle solver; no shortcut & Partly & Oracle programs, not an agent; 67 tasks checked for printed answers. \\
O.h.1--2 & Format specified; guessing minimised & Yes & Final \texttt{ANSWER:} line; 40 of 44 dollar ranges $\le\pm$\$1.00. \\
O.b.2 & Listing answers fails & Yes & Only the first number on the answer line counts. \\
O.c.1 & LLM judge validated & N/A & No judge grades; the comparison is in Section~\ref{sec:grading}. \\
R.1--2 & Open data and harness & Partly & Instructions and three examples in the paper. \\
R.3--5 & Contamination; updates; maintenance & Partly & Private keys, public scans. \\
R.6--10 & Construct; subject; flaws; impact & Yes & Sections~\ref{sec:intro}, \ref{sec:setup}, \ref{sec:qc} and~\ref{sec:limitations}. \\
R.11--12 & Significance; interpretation & Yes & Bootstrap intervals, Holm-corrected tests; Section~\ref{sec:limitations}. \\
R.13--14 & Non-AI baseline; trivial agents & No; Partly & No human baseline; no number or a known wrong answer fails. \\
\bottomrule
\end{tabular}
\end{center}

\end{document}